\documentclass[journal]{IEEEtran}
\usepackage{amssymb}
\usepackage{amsmath}
\usepackage[pdftex]{graphicx}
\usepackage{gensymb}
\usepackage[nolist]{acronym} 

\usepackage{booktabs}

\ifCLASSINFOpdf
\else
\fi
\usepackage{algorithm}
\usepackage{algpseudocode}
\newcommand{\bz}{\mathbf{z}}
\newcommand{\bx}{\mathbf{x}}
\newcommand{\bW}{\mathbf{W}}
\newcommand{\bb}{\mathbf{b}}
\newcommand{\beps}{\boldsymbol{\epsilon}}
\newcommand*\kl[2]{\mathbb{KL}[#1||#2]}

\newcommand{\eq}[1]{\begin{align*}#1\end{align*}}

\newcommand{\Expc}[2]{\mathbb{E}_{#1}[#2]}

\newcommand{\losssgvb}{\mathcal{L}_{\text{sgvb}}}

\begin{acronym}[YTB]
    \acro{SGVB}{stochastic gradient variational Bayes}
    \acro{VI}{variational inference}
    \acro{KL divergence}{Kullback-Leibler divergence}
    \acro{VAE}{variational auto-encoder}
\end{acronym}

\usepackage{todonotes}
\makeatletter
\let\t@d@\todo
\def\todo#1{\t@d@[inline]{#1}}
\makeatother

\begin{document}
%
\title{Unsupervised preprocessing for Tactile Data}
%
%
%

\author{Maximilian~Karl,
        Justin~Bayer,
        and~Patrick~van~der~Smagt
\thanks{M. Karl, J. Bayer and P. van der Smagt are with the Department
of Informatics, Technische Universit\"at M\"unchen, Munich, Germany.
 Justin Bayer is also affiliated with sensed.io UG (haftungsbeschr\"ankt), M\"unchen, Germany.
 PvdS is also with fortiss, a TUM Associate Institute.
  e-mail: karlma (at) in.tum.de, bayer (at) sensed.io.  }
}

%
%

\markboth{}
{Karl \MakeLowercase{\textit{et al.}}: Unsupervised preprocessing for Tactile Data}
%



\maketitle

\begin{abstract}
Tactile information is important for gripping, stable grasp, and in-hand manipulation, yet the complexity of tactile data prevents widespread use of such sensors. We make use of an unsupervised learning algorithm that transforms the complex tactile data into a compact, latent representation without the need to record ground truth reference data. These compact representations can either be used directly in a reinforcement learning based controller or can be used to calibrate the tactile sensor to physical quantities with only a few datapoints. We show the quality of our latent representation by predicting important features and with a simple control task.
\end{abstract}


%
\IEEEpeerreviewmaketitle

\section{Introduction}
%
%
%
%

\IEEEPARstart{T}{actile} sensors are essential for proper in-hand manipulation, and many other tasks where fingers have to grasp, hold, and handle objects. 
Yet tactile sensor data is not easy to process.  The sensors are often deformable, while the data is high-dimensional, very nonlinear, and difficult to relate to physical properties such as grip force or object shape.
Their soft nature creates highly correlated data since stimulation of the sensor array activates nearby sensor points. 

We therefore propose to use unsupervised learning techniques for transforming the tactile information into a compact, common space, independent of the physical properties of the tactile sensor.
Contrary to standard calibration procedures \cite{Karl2016ML}, these algorithms do not need any ground truth like calibration does and are able to pre-process the tactile data and transform it into a decorrelated compressed format.
This empowers their use in various control tasks as well as as measurement device for other applications.

\subsection{Related Work}

There is a reasonable body of existing literature on extracting features from tactile sensors.  In \cite{wettels_haptic_2011}, the physical quantities force vectors, curvature, and point of contact are extracted. For this supervised task ground-truth was recorded. The authors also had a first take on supervised preprocessing using ICA and PCA. They also questioned whether raw data should be used instead of tactile sensors calibrated to ground truths.

In \cite{lin_estimating_????} external stimuli like point of contact or forces and torques were applied and recorded together with the tactile data. The paper describes the use of supervised machine learning algorithms to predict those stimuli from tactile data. The results show that the sensors can measure physical quantities roughly similar to those humans can.

Such physical quantities, supervisedly extracted from tactile sensors, were used in \cite{su_use_2012} and \cite{zhe_force_????}  for control. \cite{ciobanu_tactile_2013} designed a pipeline for preprocessing tactile data from a BioTac tactile sensor. Each step in the pipeline is carefully designed, implemented and requires manual tuning. All of those applications of tactile sensors used mostly supervised methods including calibration. These type of methods feature several problems which are discussed in the following.

Another problem with tactile sensors is that their sensor information is the result of several external modalities. \cite{synergistic} recorded micro vibrations from sliding over silk, suede and sandpaper surfaces and showed that different force levels changes the spectrum significantly. The two informations surface type and force can therefore not be separated easily. This shows that learning only the mappings from raw tactile data to those ground truths in a supervised way will lead too a lot of loss of information.

\subsection{Unsupervised Learning}

Calibrating tactile sensors in this supervised way poses several problems.
The procedure of recording the labelled data needs to be redone for each individual sensor and task and is very time-consuming.
Another problem lies in the choice of ground truth applied to the sensor as one might miss important features if they are not carefully chosen.
Unsupervised learning algorithms might be able to solve both problems.
It does not require any ground truth and can find intrinsic structures in the tactile data itself.
We are looking for models creating a mathematically compact or dimensionally-reduced representation of the data.
This representation is more suited for control algorithms than the raw, high-dimensional, nonlinear tactile data.
In the case where a calibration to physical values is still required, only a small number of labelled data is needed to find the relation between the compressed representation and the physical ground truth.

Our approach is based on graphical models with latent variables.
These latent variables represent the compact representation and can be found by probabilistic inference.


\section{Variational Auto-Encoder}

\subsection{Linear latent models}

Latent variable models offer a mathematically well-founded framework for the extraction of features from data.
Sparse coding, Independent component analysis (ICA) and principal component analysis (PCA) are linear variants thereof and can be formulated as solutions to the optimisation of the marginal likelihood of the data $\bx$.
The observations are conditioned on the latent variables $\bz$, which are subsequently marginalised out:
\eq{
    p(\bx) = \int_\bz p(\bx\mid\bz)\,p(\bz) \,d\bz.
}
When the aforementioned methods are cast into this framework, approximations are typically used, such as \emph{maximum a posteriori} inference for the latent variables in sparse coding.

A drastic limitation of linear latent variable models is the fact that $\bx$ depends only linearly on $\bz$, e.g.\ $\bx = \bz \bW + \bb + \beps$, where $\bW$ is a matrix, $\bb$ a vector offset and $\beps$ i.i.d.\ noise variables.
Arguably, these models are unable to perform nonlinear transformations of the data. 
In settings where complex sensors (e.g., an array of spherically arranged sensors) or nonlinear relationships between physics and sensor are involved, this is clearly not sufficient.
Representing the observations $\bx$ through a non-linear transformation of the latent variables $\bz$, i.e.\ $\bx = f(\bz)$ is appealing but challenging. 
We will review a method for that case in the next section.

\subsection{Stochastic Gradient Variational Bayes}

Recently, an efficient method to estimate such nonlinear functions called \ac{SGVB} has been proposed \cite{kingma_stochastic_2013,rezende_stochastic_2014}.
In SGVB, the latent variables have to be continuous random variables, and are typically chosen as zero-centred Gaussians with  identity covariance matrix.

 \ac{VI} lies at the basis of SGVB.  In VI, probability distributions are approximated by finding the closest member of a restricted family of distributions by means of optimisation.
It turns out that in the case of latent variable models, a tractable objective function can be found: the variational upper bound on the negative log-likelihood. The derivation can be summarised as follows:
\eq{
-\log p(\bx) 
    &= \log \int_\bz p(\bx\mid\bz) p(\bz) d\bz \\
    &= \log \int_\bz \frac{q(\bz)}{q(\bz)} p(\bx\mid\bz) p(\bz) d\bz \\
    &\textrm{using Jensen's inequality:}\\
    &\geq \int_\bz q(\bz) \log  \frac{p(\bx\mid\bz) p(\bz)}{q(\bz)}  d\bz \\
    &=-\Expc{\bz \sim q(\bz)}{\log p(\bx\mid\bz)} + \kl{q(\bz)}{p(\bz)} \\
    &=: \mathcal{L}.
}
Herre, $\kl qp$ is the Kullback-Leibler divergence between two probability densities, and expresses how different they are.

\ac{SGVB} takes this one step further and implements $q$ as a neural network conditioned on the input, i.e., $q(\bz|\bx; \theta)$ where $\theta$ is a set of weights of the neural network.
The neural network with input $\bx$ and weights $\theta$ would generate a mean $\mu(\bx)$ and standard deviation $\sigma(\bx)$ for modelling $\bz$ as a Gaussian distribution.
The likelihood is also implemented as a neural network $p(\bx|\bz; \theta)$; hence the name \ac{VAE}.
The objective loss function then is
\eq{
    \losssgvb &:=-\Expc{\bz \sim q(\bz|\bx; \theta)}{\log p(\bx|\bz; \theta)} + \kl{q(\bz|\bx; \theta)}{p(\bz)}.
}
Given a solution to this problem, we will obtain an efficient mean to evaluate $p(\bx|\bz)$ with a simple forward pass through a neural network.
Further, it can be shown that $q(\bz|\bx)$ will be close to the true but intractable posterior $p(\bz|\bx)$.
SVGB therefore poses a mean to efficiently extract latent variables $\bz$ from observations $\bx$.

Obtaining a solution can be done by stochastic gradient descent: by sampling from $q$, we can approximate the expectation in the loss.
In the case of both $q(\bz|\bx)$ and $p(\bz)$ being diagonal Gaussians, the KL-divergence can be evaluated efficiently in closed form.

\section{Setup}

\subsection{Tactile Sensors}

We used two types of tactile sensors. The BioTac Sensor \cite{biotacmanual, fishel_design_2012, fishel_sensing_2012} and the tactile sensor from the iCub robot \cite{schmitz_tactile_2010}. The BioTac sensor consists of a soft, liquid-filled silicone membrane over a hard core while the iCub is comparatively stiffer with a soft but very thin coating. Both sensor also differ in the measurement principle with the BioTac measuring the electrical impedance of the liquid and the iCub measuring the capacity of its coating. The BioTac sensor also features sensors measuring the pressure, vibrations and temperature of the liquid but these values were ignored in the following experiments. Further details on the amount and types of sensors can be found in Table: \ref{tab:sensor_comparison}.

\begin{table}[h]
    \caption{Sensor comparison}
    \label{tab:sensor_comparison}
    \centering
    \begin{tabular}{l|c|c}
        & BioTac       & iCub       \\ \midrule
        \multicolumn{1}{l|}{\# taxels}        & 19           & 12         \\ 
        \multicolumn{1}{l|}{measurement principle}           & resistive    & capacitive \\ 
        \multicolumn{1}{l|}{DC Pressure Range}    & 0--100\,kPa    & --          \\ 
        \multicolumn{1}{l|}{DC Temp. Range}       & 0--75$\degree$     & --         \\ 
        \multicolumn{1}{l|}{AC Pressure spectrum} & 10--1040\,Hz   & --          \\ 
        \multicolumn{1}{l|}{AC Temp. spectrum}    & 0.45--22.6\,Hz & --          \\ 
    \end{tabular}
\end{table}

\begin{figure}
    \includegraphics[width=\linewidth,]{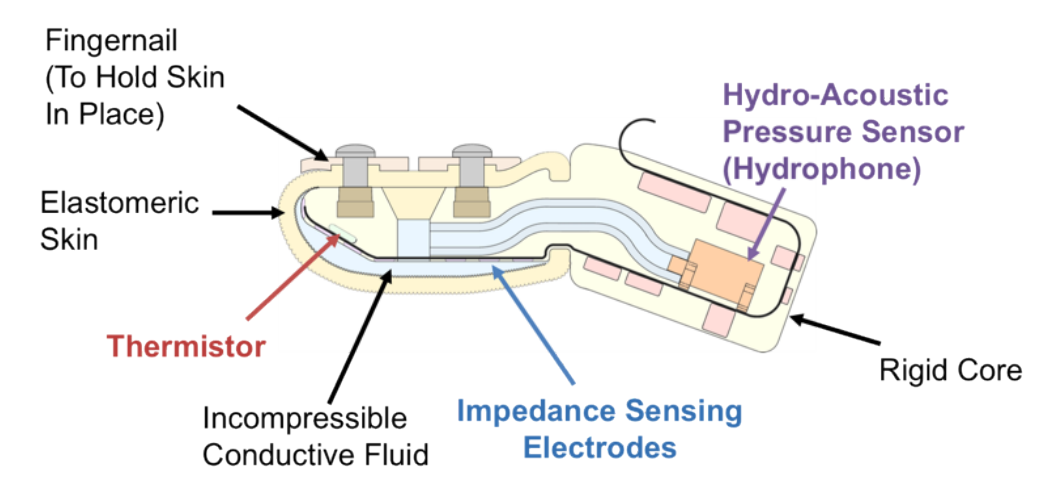}
    \caption{Sideways cross-section of BioTac \cite{biotacmanual} \cite{fishel_design_2012}}
    \label{fig:biotac_side}
\end{figure}

\begin{figure}
    \includegraphics[width=\linewidth,]{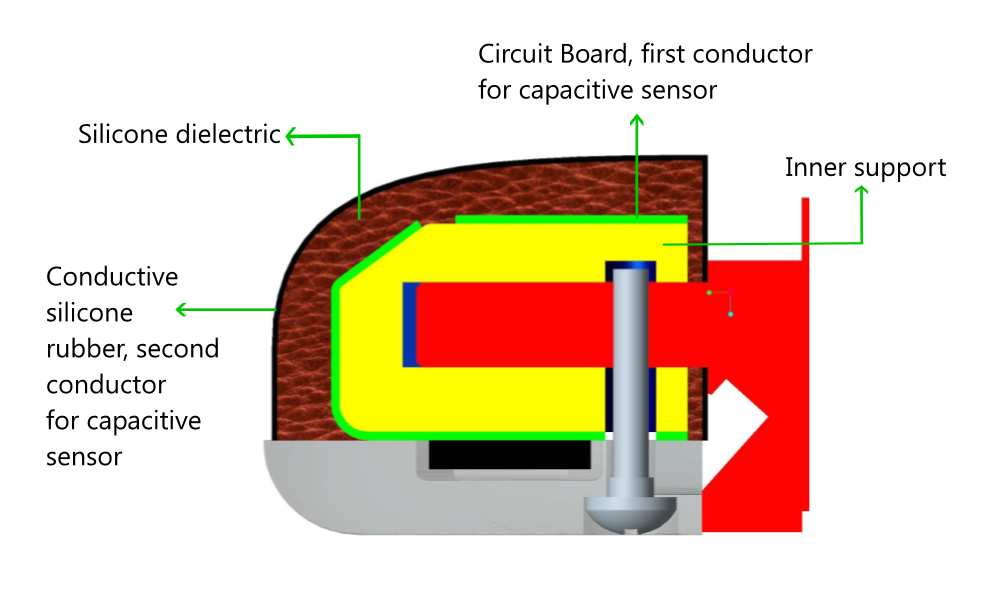}
    \caption{Sideways cross section of the iCub sensor \cite{schmitz_tactile_2010}.}
    \label{fig:icub_side}
\end{figure}

\subsection{Test Bed}

For verifying our unsupervised learning methods we performed several experiments with different stimuli and recorded the tactile data.
These stimuli include force, shore hardness, surface angles and curvatures.
They were chosen such that they represent important information which are relevant for grip and manipulation.
To verify the representation in the latent space learned by the neural networks, we also recorded ground truth during our data set measurement.  
These ground-truth data are not used in the neural network training process.

For measuring accurate and repeatable datasets, a small 3-DoF robot with an additional linear actor was set up to fit our needs.

\begin{figure*}
    \includegraphics[width=\linewidth,]{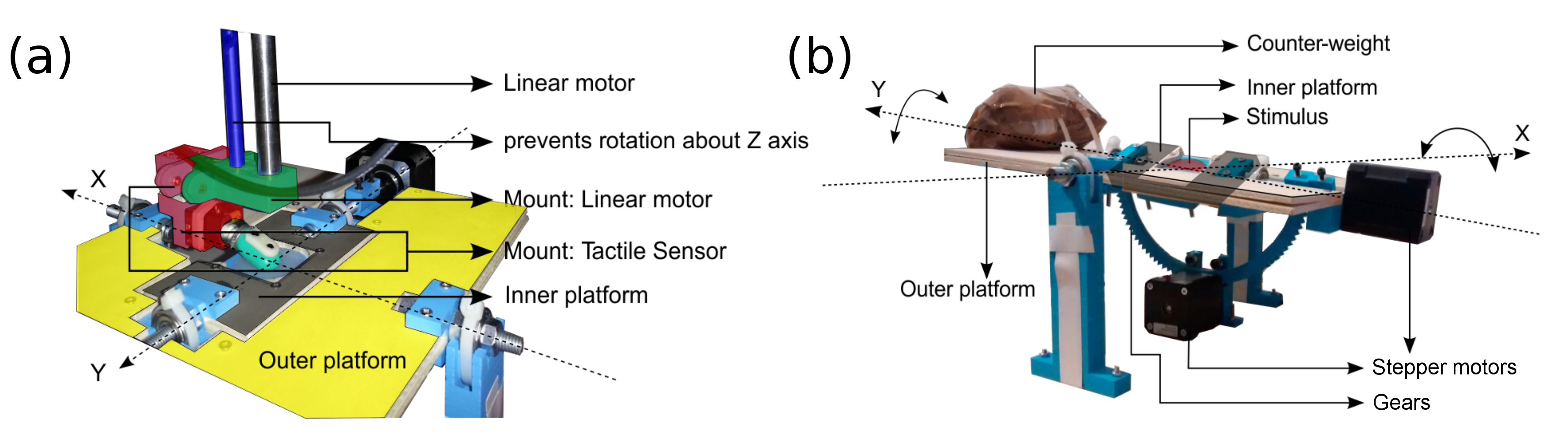}
    \caption{Robot setup with gimbal axis for dataset recording (a) End effector of robot touching sample inside gimbal platform. (b) Gimbal platform in detail.}
    \label{fig:robotsetup}
\end{figure*}

It was equipped with a mount for holding different types of tactile sensors as well as a force-torque sensor (ATI Nano 17) to control the force applied with a linear actor.
Additionally, a gimbal platform was added for measuring materials at different angles.
The robotic setup is shown in Fig.~\ref{fig:robotsetup}.
The two degrees of freedom of this platform were actuated by computer-controlled stepper motors.
The centre platform of the gimbal axis is replaceable, allowing us to evaluate different materials.
The electronics of the robot are connected to a PC using an FPGA PCI-Card (Mesa 5i25) to support real-time control using Matlab Simulink. The operating system for this desktop computer is the Matlab Simulink Real-Time XPC operating system. 
All sensors are either directly connected or connected through a microprocessor to this FPGA card.
This ensures proper time synchronisation and a constant delay of 30\,ms between sampled data points.

\subsection{Datasets}

\begin{figure*}
    \includegraphics[width=\linewidth,]{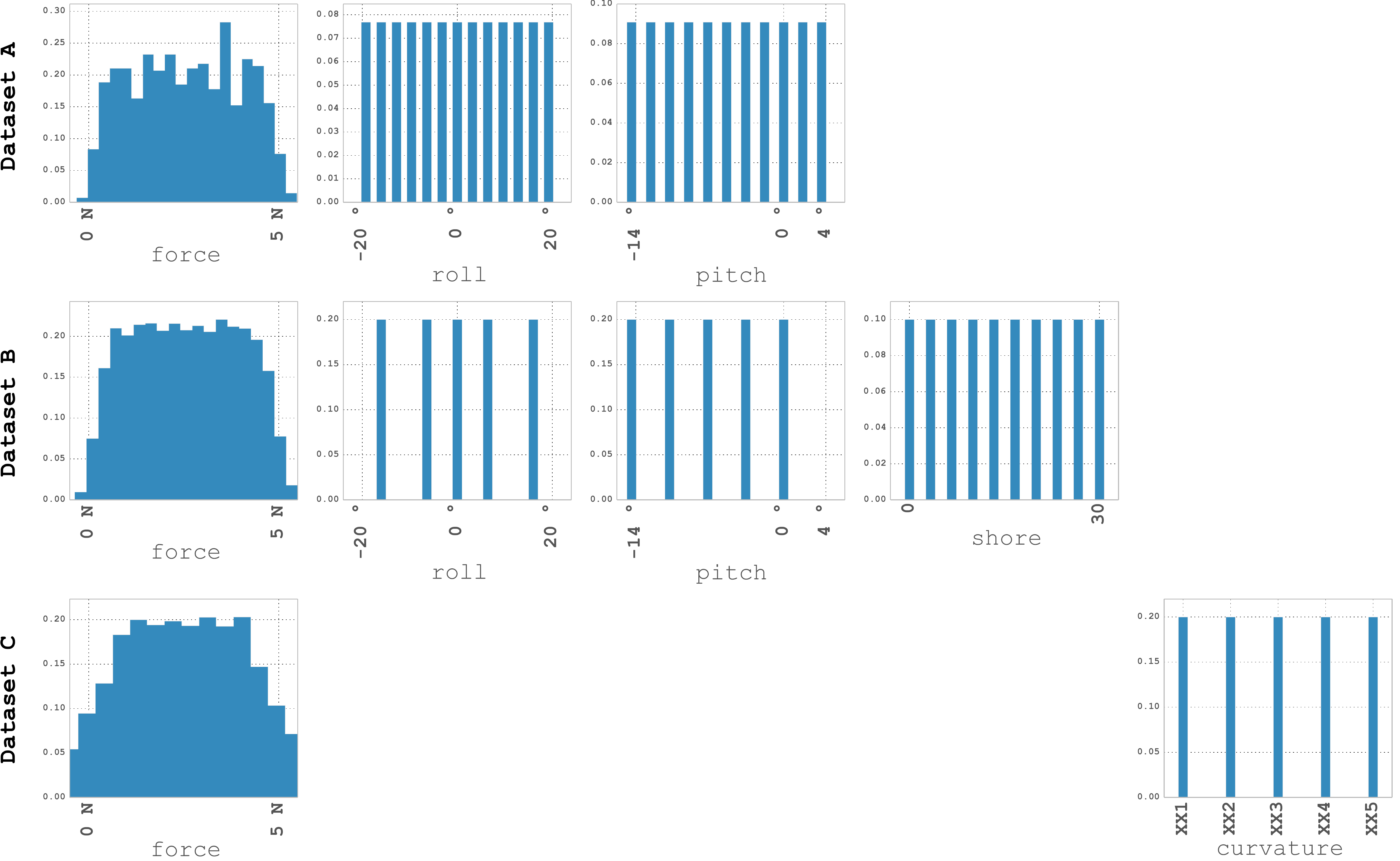}
    \caption{Histograms of the reproducible dataset recorded with the modified robot platform. (Dataset A) Surface angle dataset. (Dataset B) Surface angle dataset including changing shore hardness. (Dataset C) Curvatures dataset
}
    \label{fig:datasets}
\end{figure*}

The external stimuli were chosen to include shore hardness, surface normal and curvature.

\paragraph{Surface angle estimation.}
The angular dataset was created by setting the gimbal axis to a fixed angle followed by linear increasing the force of the tactile sensor pressing against the gimbal platform.
After reaching 5\,N, the force was decreased at the same speed to capture possible hysteresis effects.
The angles of the gimbal platform are then changed and the application of the tactile sensor is repeated. As material in the gimbal axis centre a flat plastic surface was used.
The angle ranges from $-19$\degree~to 19\degree~in the roll direction and from $-3.6$\degree~to 18\degree~in the pitch direction.

\paragraph{Shore hardness estimation.}
The material samples for shore hardness were created using a two component silicone (Smooth On Ecoflex) and then calibrated using a Shore A measurement tool.
Using this technique we managed to get a uniform distribution between 0 and 30 Shore A.
These material samples are shaped as cylindrical plates with a diameter of 4\,cm and a height of 1\,cm.
A 3-D printed plastic holder for receiving these silicone plates was mounted inside the centre of the gimbal axis.
The dataset was not only recorded for a planar angle, but also measured at different angles using the gimbal platform.
See Fig.~\ref{fig:datasets}, Dataset B: each Shore hardness shown in the last plot in that row is recorded together with the shown variation of angles.

\begin{figure}
   \centerline{ \includegraphics[width=0.7\linewidth,]{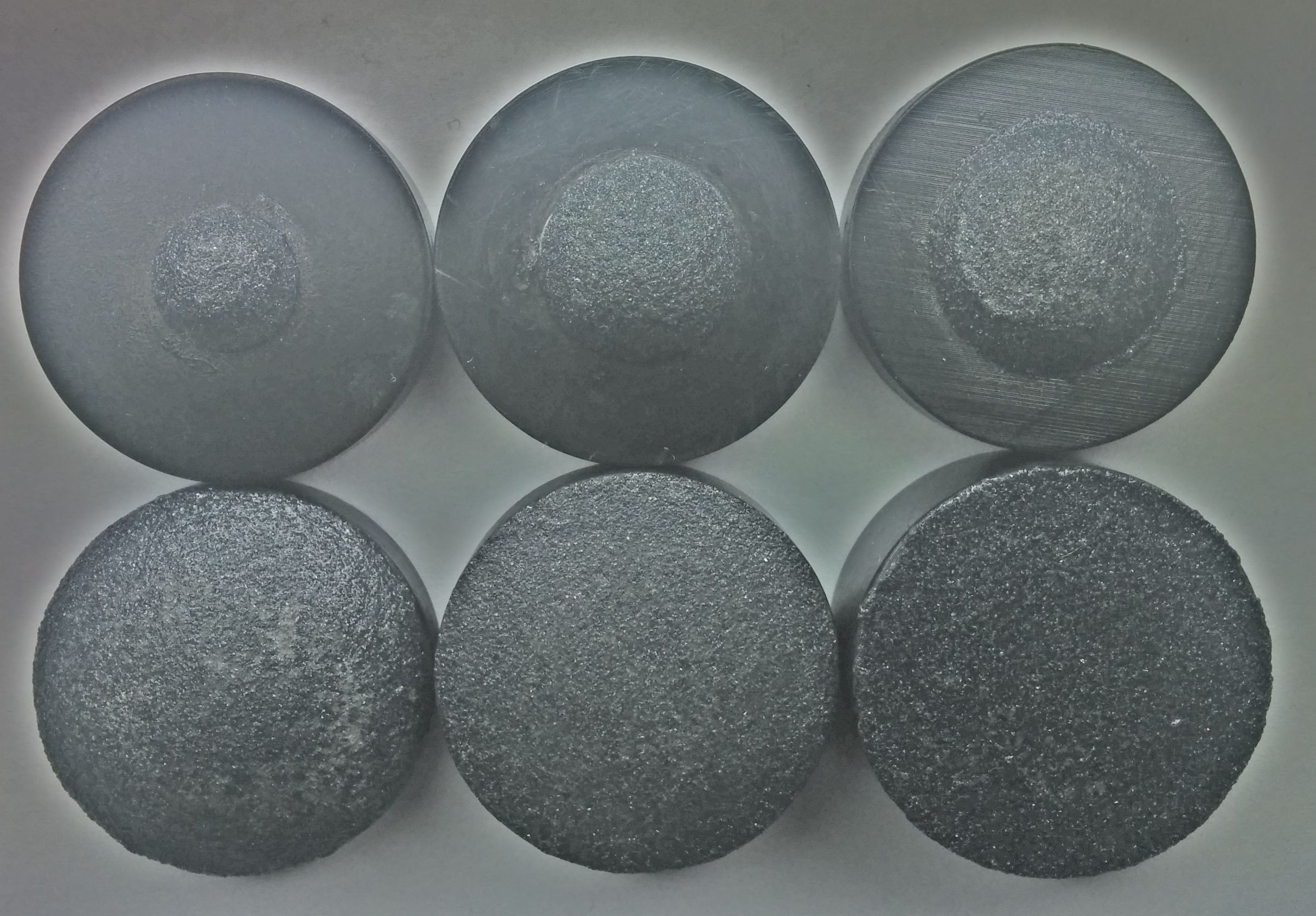}}
    \caption{Curved objects used for the curvature dataset. Radii top row: 5\,mm, 7.5\,mm, 10\,mm. Bottom row: 20\,mm, 40\,mm, flat. Picture postprocessed for better examination.}
    \label{fig:curvatures}
\end{figure}

\emph{Curvature estimation} The curvature dataset is obtained from six different spherical curvatures. With five radii ranging from 40\,mm to 5\,mm.
The force was controlled to be uniformly distributed between 0--5\,N for both
sensors. These curvature samples are shown in Fig.~\ref{fig:curvatures}.

An overview over these datasets, their attribute ranges and distributions are shown in Fig.~\ref{fig:datasets}.
The rows represent the different Datasets and the columns different attributes.
Note that every dataset consists of all possible combinations of its shown attribute values.

\section{Results}

For all experiments and different tactile sensors we used the following common VAE network configuration.
The generative model is defined as
\[
  \log p(\bx\mid\bz) = \log \mathcal{N}(\bx; \mu(\bz), \sigma^2 I),
\]
where $\sigma$ is a constant in all dimensions of $\bx$ and $\mu(\bz)$ is a neural network with two layers each 512 elements wide. $\sigma$ is part of the parameters and thus subject to the optimisation. 

The recognition model is defined as
\[
  \log p(\bz\mid\bx) = \log \mathcal{N}(\bz; \mu(\bx), \sigma(\bx)^2 I),
\]
where a neural network model outputs $\log \sigma(\bx)^2$ and $\mu(\bx)$ as a concatenated vector.
The recognition model neural network has the same size, transfer functions and number of hidden layers as the generative model.
We used the identity function for all output transfer function of the neural networks.

The only differences between the networks for BioTac and iCub VAE networks lies in the transfer functions and optimiser used. For the BioTac we used sigmoid transfer functions for all hidden layers in both recognition and generative model networks whereas the network for iCub data used rectifier transfer functions for all hidden layers.
The optimiser used for the iCub VAE was adadelta\cite{adadelta} with step rate 0.1, and for the BioTac VAE rmsprop\cite{rmsprop} with step-rate 0.001.

For showing the differences between preprocessed data and raw data and verifying that all information is still encoded in the features after applying the VAE we evaluated the prediction quality on the measured ground truth.
We used linear regression, decision tree regression and multilayer perceptrons to evaluate the quality of the latent space.

\begin{table}
\caption{Regression results on raw and unsupervised preprocessed data for BioTac sensor on surface dataset. Best result for each pair in bold.}
\label{tab:regression_biotac_surface}
\begin{tabular}{ l c c | c c }
& \multicolumn{2}{c}{Linear Regression}  & \multicolumn{2}{c}{Decision Trees}  \\
& raw & latent & raw & latent \\
Force [N] & 0.32 & \textbf{0.27} & \textbf{0.34} & 0.36 \\
Pitch [\degree] & 4.02 & \textbf{3.18} & \textbf{3.69} & 3.70 \\
Roll [\degree] & 7.03 & \textbf{4.79} & 6.90 & \textbf{5.96} \\
\end{tabular}
\end{table}

\begin{table}
\caption{Regression results on raw and unsupervised preprocessed data for iCub sensor on surface dataset. Best result for each pair in bold.}
\label{tab:regression_icub_surface}
\begin{tabular}{ l c c | c c }
& \multicolumn{2}{c}{Linear Regression}  & \multicolumn{2}{c}{Decision Tree Regression}  \\
& raw & latent & raw & latent \\
Force [N] & 0.86 & \textbf{0.75} & 0.87 & \textbf{0.81} \\
Pitch [\degree] & 4.34 & \textbf{2.92} & 3.34 & \textbf{2.80} \\
Roll [\degree] & 5.23 & \textbf{3.86} & \textbf{3.73} & 4.84 \\
\end{tabular}
\end{table}

\begin{table}
\caption{Regression results on raw and unsupervised preprocessed data for BioTac sensor on shore dataset. Best result for each pair in bold.}
\label{tab:regression_biotac_shore}
\begin{tabular}{ l c c | c c }
& \multicolumn{2}{c}{Linear Regression}  & \multicolumn{2}{c}{Decision Trees}  \\
& raw & latent & raw & latent \\
Force [N] & \textbf{0.30} & 0.34 & \textbf{0.33} & 0.37 \\
Pitch [\degree] & 3.45 & \textbf{2.93} & \textbf{3.47} & 3.62 \\
Roll [\degree] & 6.19 & \textbf{4.48} & \textbf{6.19} & 6.45 \\
Shore [Shore A] & 2.07 & \textbf{1.94} & 2.21 & 2.21 \\
\end{tabular}
\end{table}

\begin{table}
\caption{Regression results on raw and unsupervised preprocessed data for iCub sensor on shore dataset. Best result for each pair in bold.}
\label{tab:regression_icub_shore}
\begin{tabular}{ l c c | c c }
& \multicolumn{2}{c}{Linear Regression}  & \multicolumn{2}{c}{Decision Tree Regression}  \\
& raw & latent & raw & latent \\
Force [N] & 0.56 & \textbf{0.49} & 0.68 & \textbf{0.67} \\
Pitch [\degree] & 2.45 & \textbf{1.58} & 2.50 & \textbf{2.09} \\
Roll [\degree] & 4.39 & \textbf{2.47} & 3.53 & \textbf{3.47} \\
Shore [Shore A] & 1.99 & \textbf{1.37} & 2.35 & \textbf{1.76} \\
\end{tabular}
\end{table}

\subsection{Linear Regression}

\begin{figure}[!t]
\centering
\begin{tabular}{ c c c }
           & BioTac & iCub \\
    Raw    &  \begin{minipage}{0.15\textwidth} \includegraphics[width=\textwidth]{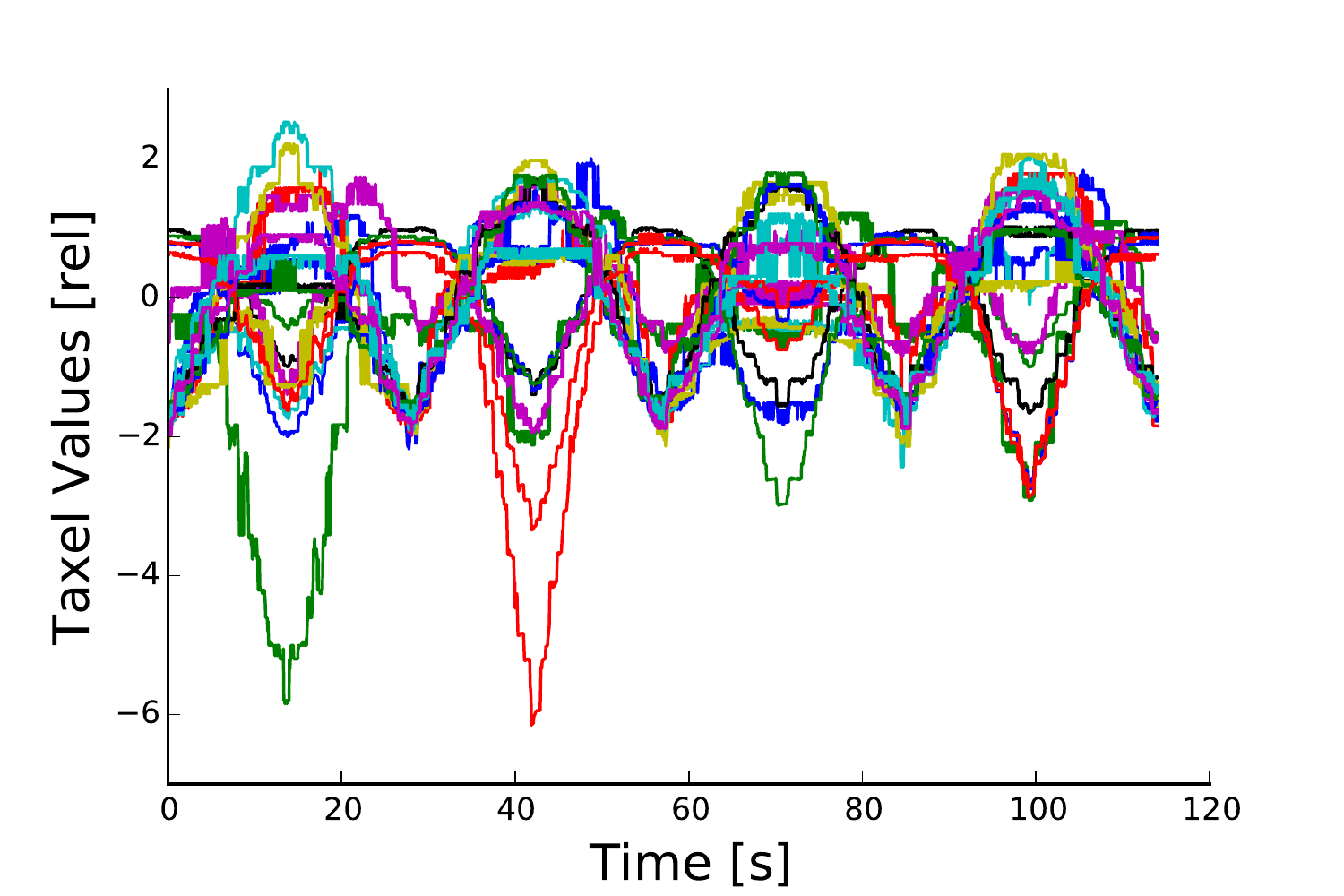}\end{minipage} & \begin{minipage}{0.15\textwidth} \includegraphics[width=\textwidth]{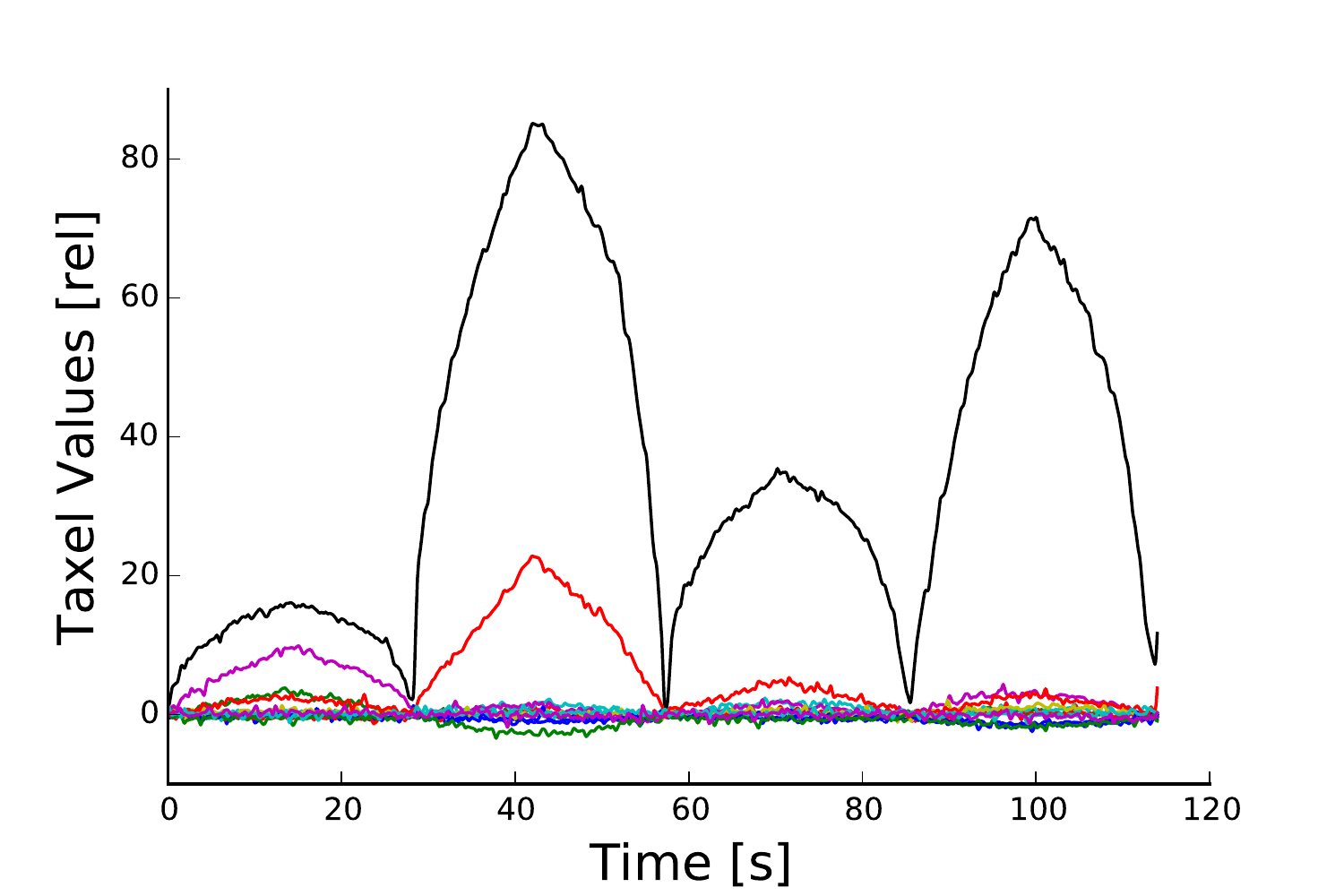}\end{minipage}\\
    Latent & \begin{minipage}{0.15\textwidth} \includegraphics[width=\textwidth]{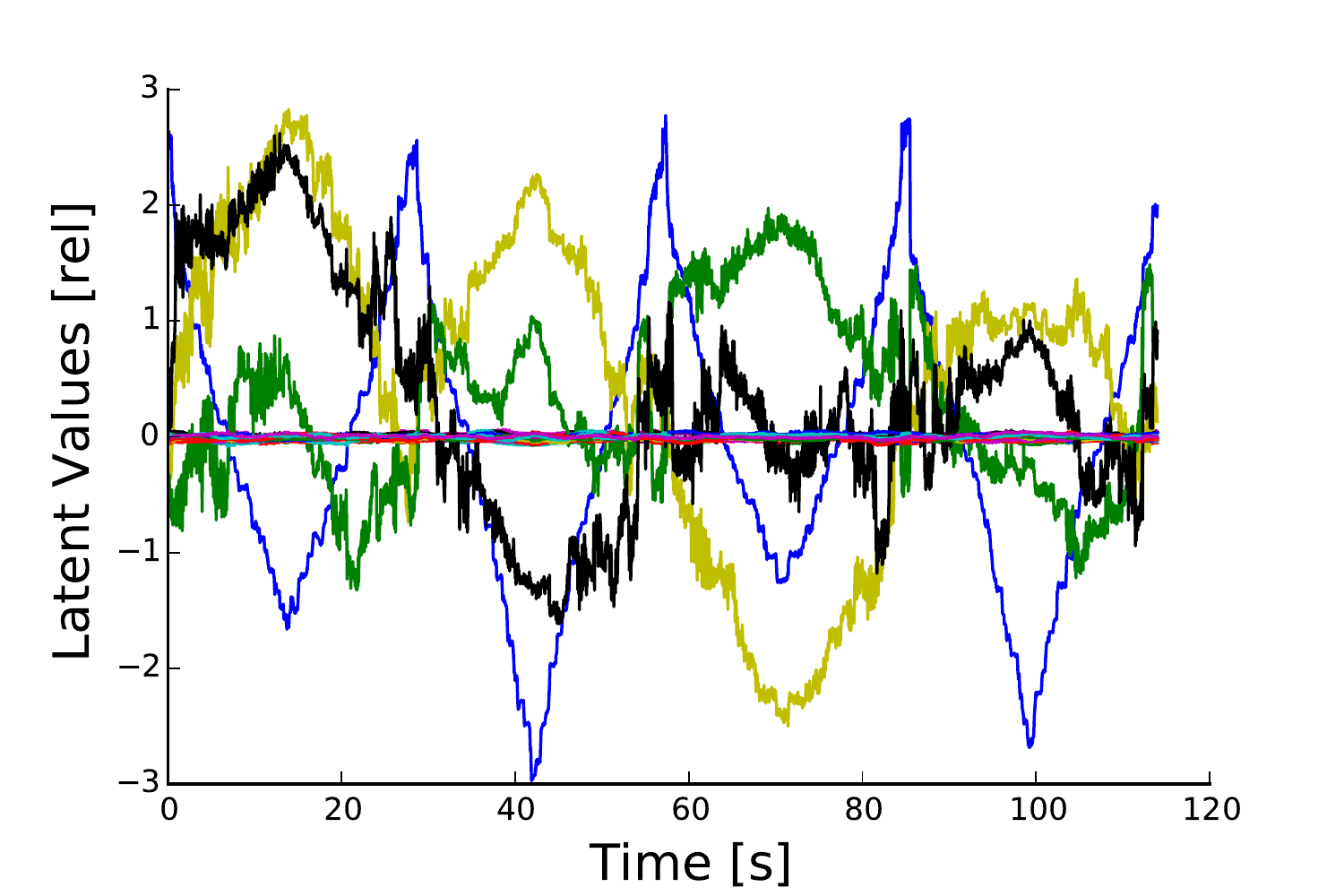}\end{minipage} & \begin{minipage}{0.15\textwidth} \includegraphics[width=\textwidth]{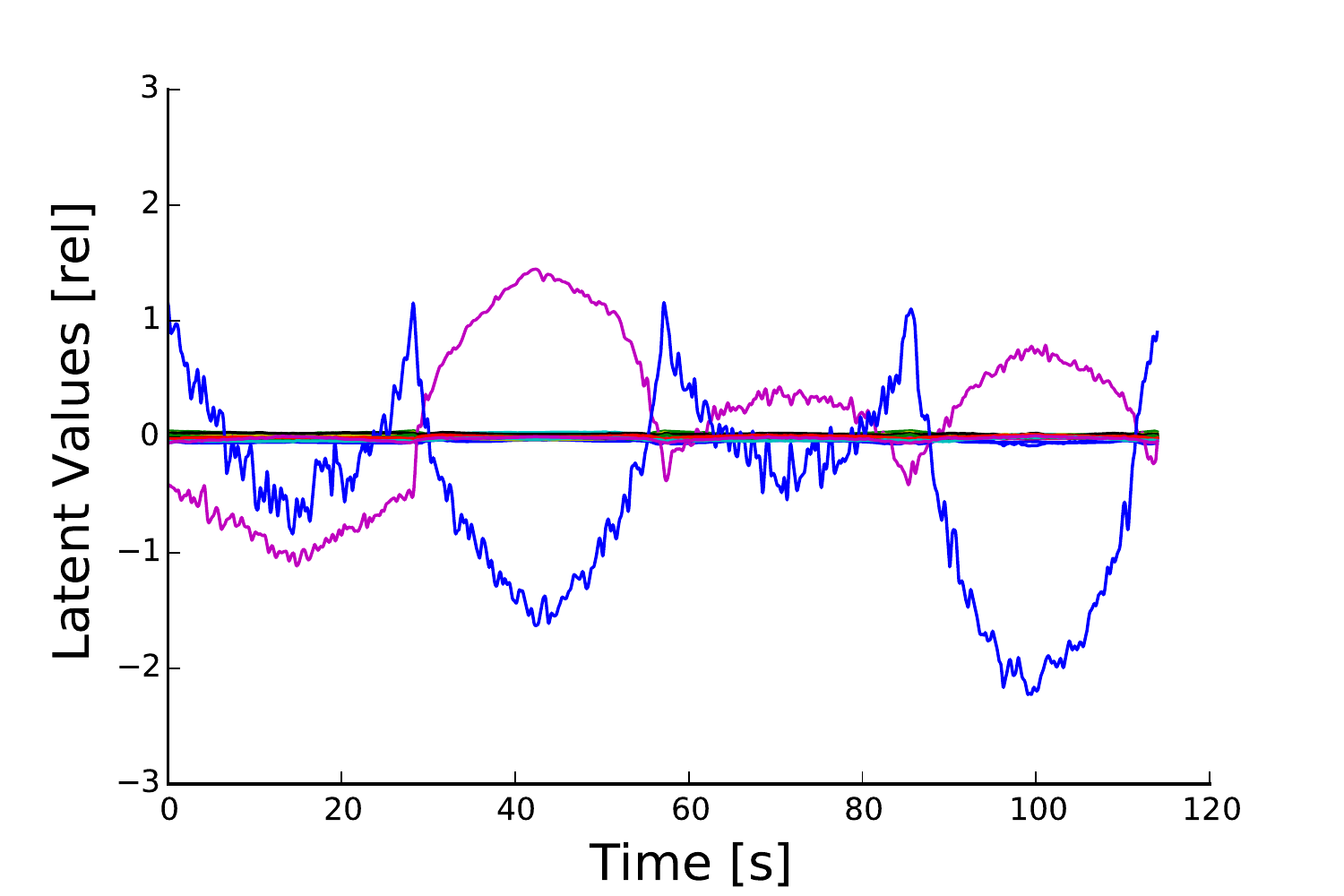}\end{minipage} \\
\end{tabular}
\caption{Raw tactile responses for BioTac (top) and iCub (bottom). Both sensor received the same stimulation of a linear increasing and decreasing force from 0\,N to 5\,N. Both sensors show huge differences in non-linearity and number of responding channels. These difference are due to differences in the material and measurement principle. }
\label{fig:sensor_raw}
\end{figure}

We noticed a large difference in the results between raw and pre-processed data when using linear regression.
As seen in Tables~\ref{tab:regression_biotac_surface}, \ref{tab:regression_icub_surface}, \ref{tab:regression_biotac_shore} and \ref{tab:regression_icub_shore} ,linear regression on the Variational Auto-Encoder pre-processed data almost always outperforms the results on raw data for both surface and shore dataset.
The inferior quality of linear regression on raw data can be explained by the highly nonlinear, sparse representation of the stimuli in the raw tactile data as shown in Fig.~\ref{fig:sensor_raw}.
The plots show the recorded raw tactile data for both sensors while applying the same force profile for both sensors.
The force profile consists of linearly increasing the force for approximately 15 seconds from 0\,N to 5\,N and then decreasing it again at the same speed.

The BioTac sensor shows a highly non-linear relation to the applied force and almost all of the 19 taxels respond to the force change.
The iCub sensor shows a different reaction with less nonlinearity and only a few taxels active at the same time.
Both nonlinearity and the selective activation of taxels are disadvantageous for algorithms like linear regression.
The improved results after preprocessing make sense as the VAE is factorising important features into individual latent variables which helps the linear regression to predict the ground truth.

\subsection{Decision Tree Regression}

Smaller differences between results can be seen in the case of Decision Tree Regression.
The Decision Trees can represent more complex relations than linear transformations and can therefore incorporate the non-linear transformations which would otherwise be applied by the Variational Auto-Encoder.

\subsection{Linear Classification on Curvature}

\begin{figure}
\centering

\begin{tabular}{ c c }
    raw sensor space & latent sensor space \\
      \includegraphics[width=1.5in]{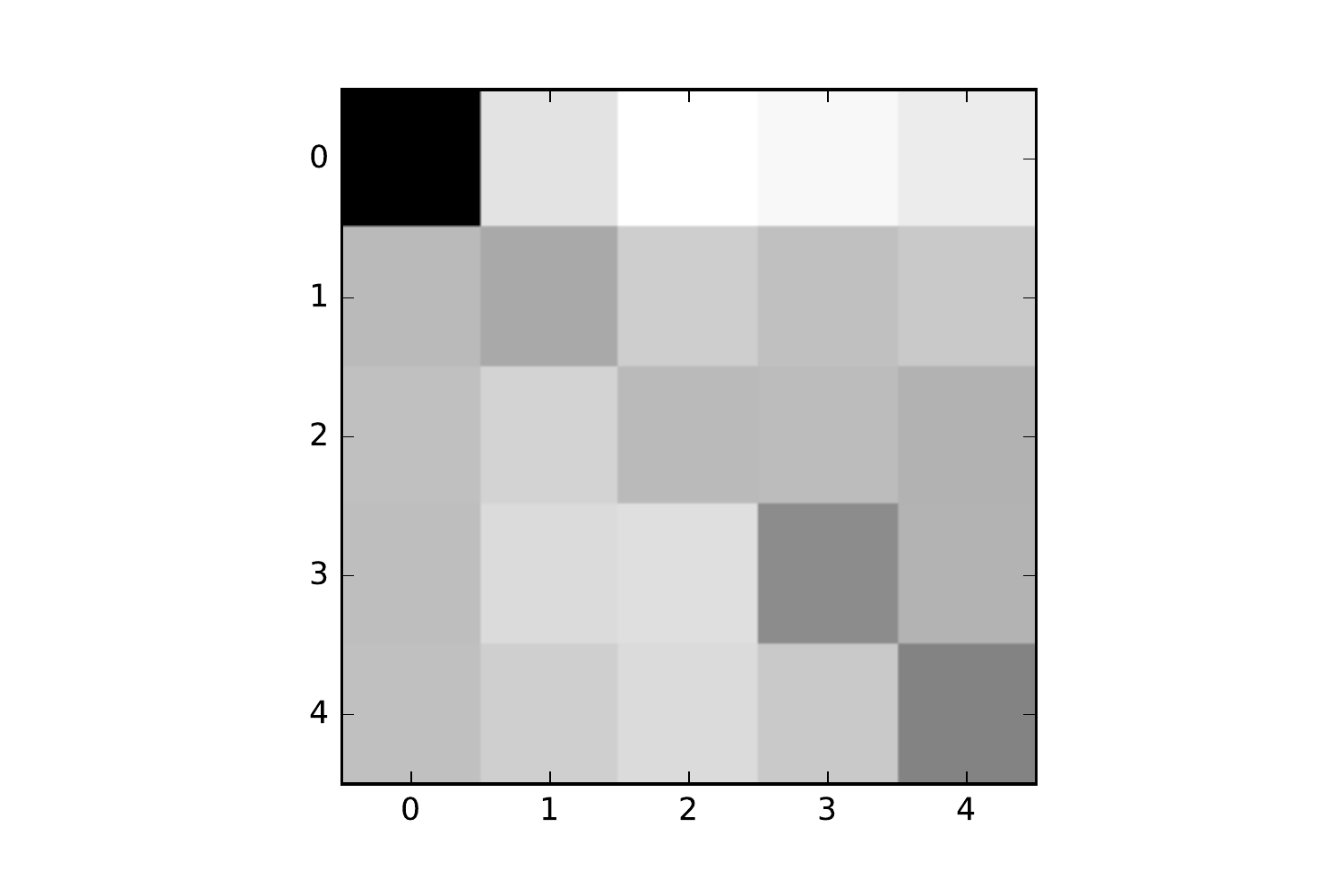} & \includegraphics[width=1.5in]{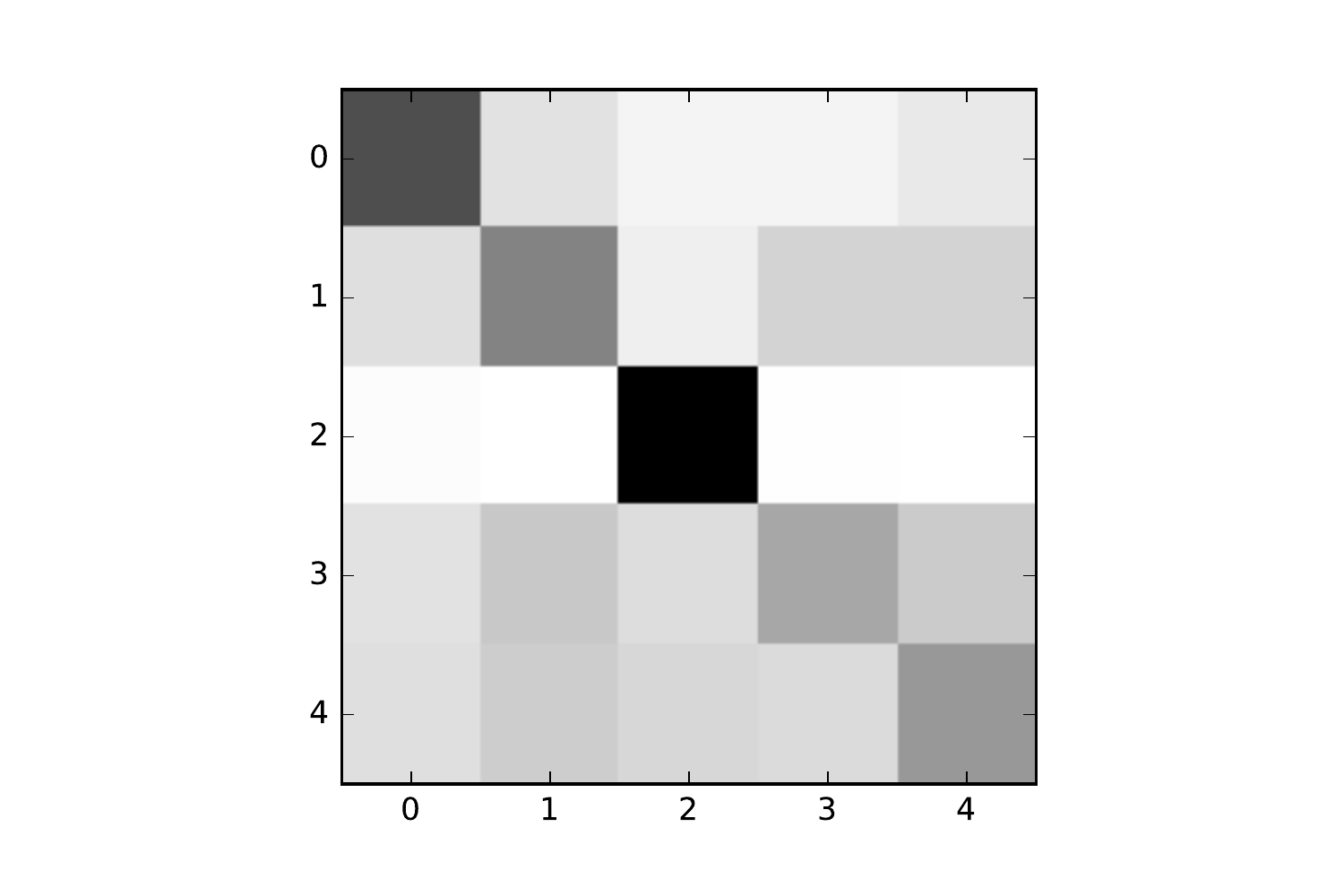}
\end{tabular}

\caption{Linear classification on curvature data for the iCub sensor. (left) confusion matrix for raw sensor data (right) confusion matrix for preprocessed sensor data.}
\label{fig:curvatureicub}
\end{figure}

\begin{figure}
\centering

\begin{tabular}{ c c }
    raw sensor space & latent sensor space \\
      \includegraphics[width=1.5in]{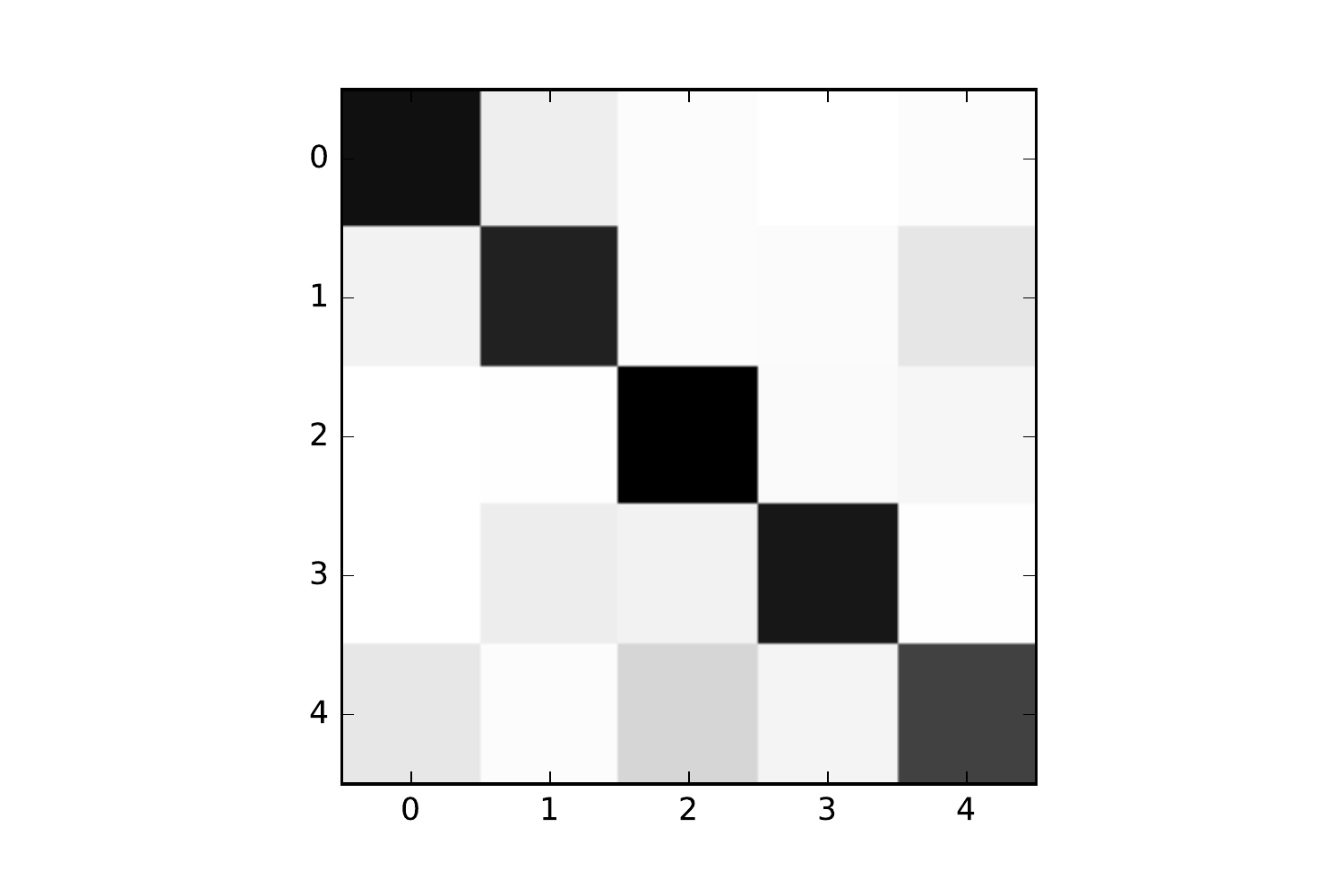} & \includegraphics[width=1.5in]{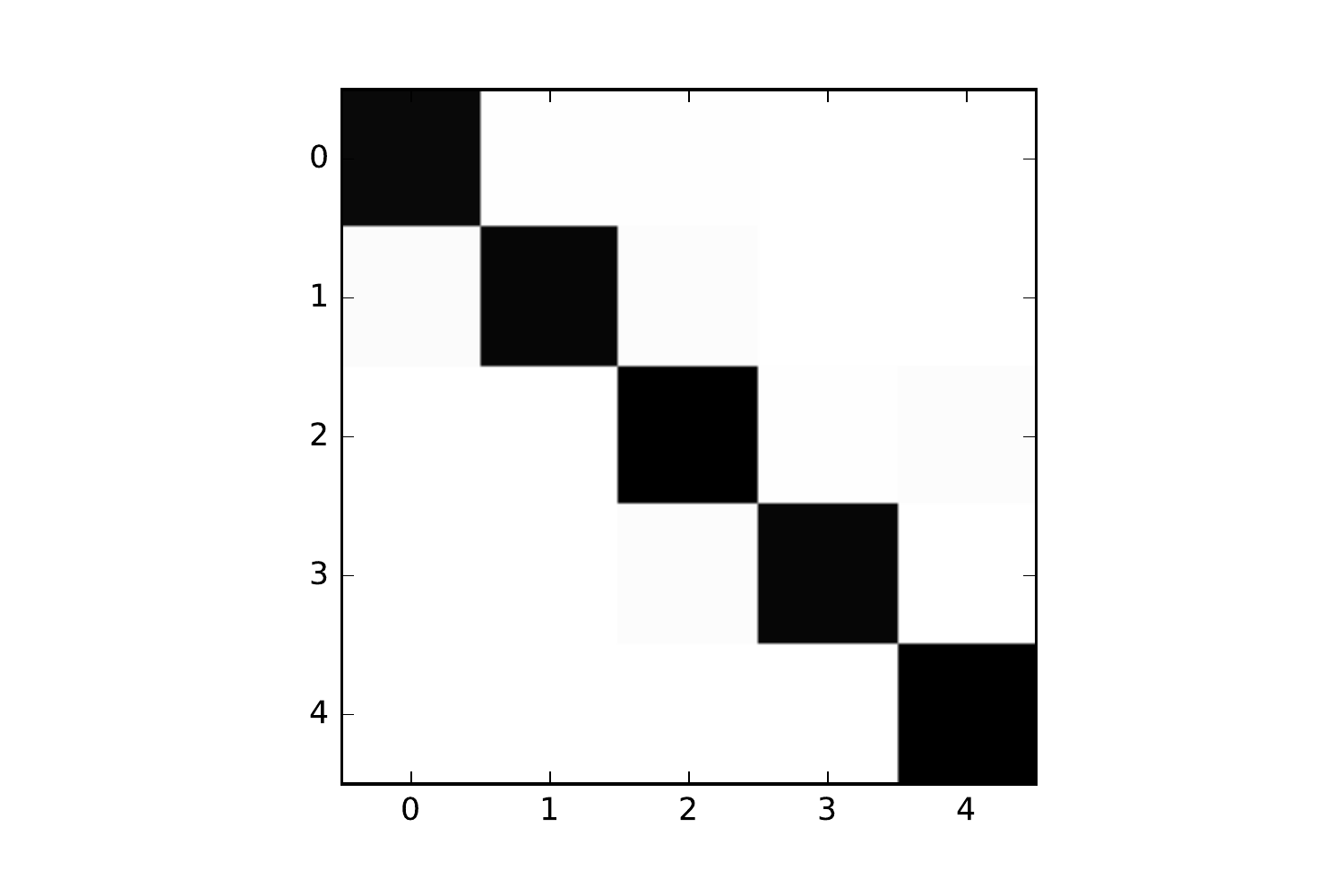}
\end{tabular}

\caption{Linear classification on curvature data for the BioTac sensor. (left) confusion matrix for raw sensor data (right) confusion matrix for preprocessed sensor data.}
\label{fig:curvaturebiotac}
\end{figure}

The curvature dataset consisted of data from five discrete curvature samples.
We used linear classification for evaluating the predictive capabilities of the latent and the raw sensor space.
Results are shown in Fig.~\ref{fig:curvatureicub} and Fig.~\ref{fig:curvaturebiotac}.
The left plot shows the confusion matrix for the raw sensor space and the right plot shows the result for the latent sensor space.
A clear diagonal represents a good prediction result.
We see that a better result for curvature classification can be achieved by transforming the raw sensor data into a more compact latent state.

\subsection{Evaluation of latent space}

We saw that all information about the applied stimulation is still present after preprocessing using the VAE and the tactile information is now in a representation suited for linear regression or linear controller.
In a real-world robotics setup with several tactile sensors we will not have any access to the real ground truth unless we perform a tedious calibration.
We would therefore benefit from a method capable of preprocessing tactile data to a similar format like a sensor calibrated to physical quantities.

\begin{figure}
\centering

\begin{tabular}{ c c }
		BioTac & iCub \\
      \includegraphics[width=1.5in]{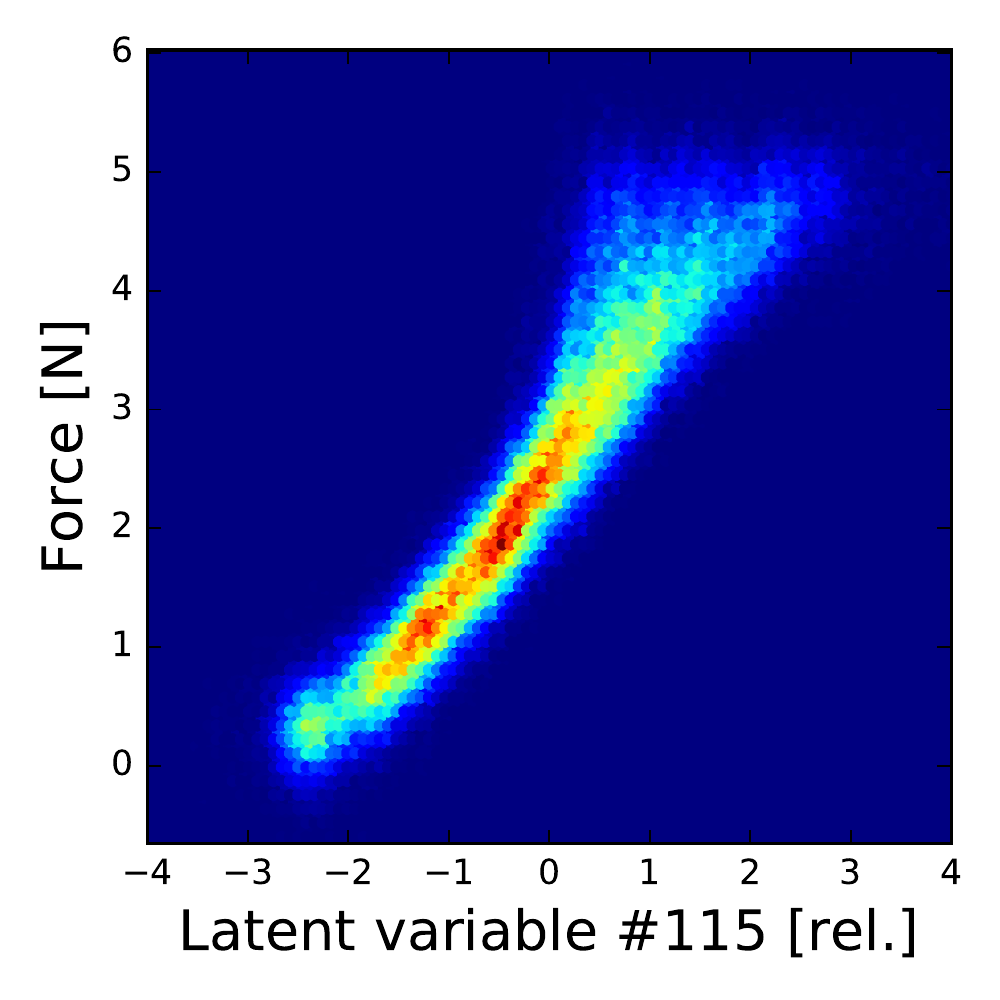} &
      \includegraphics[width=1.5in]{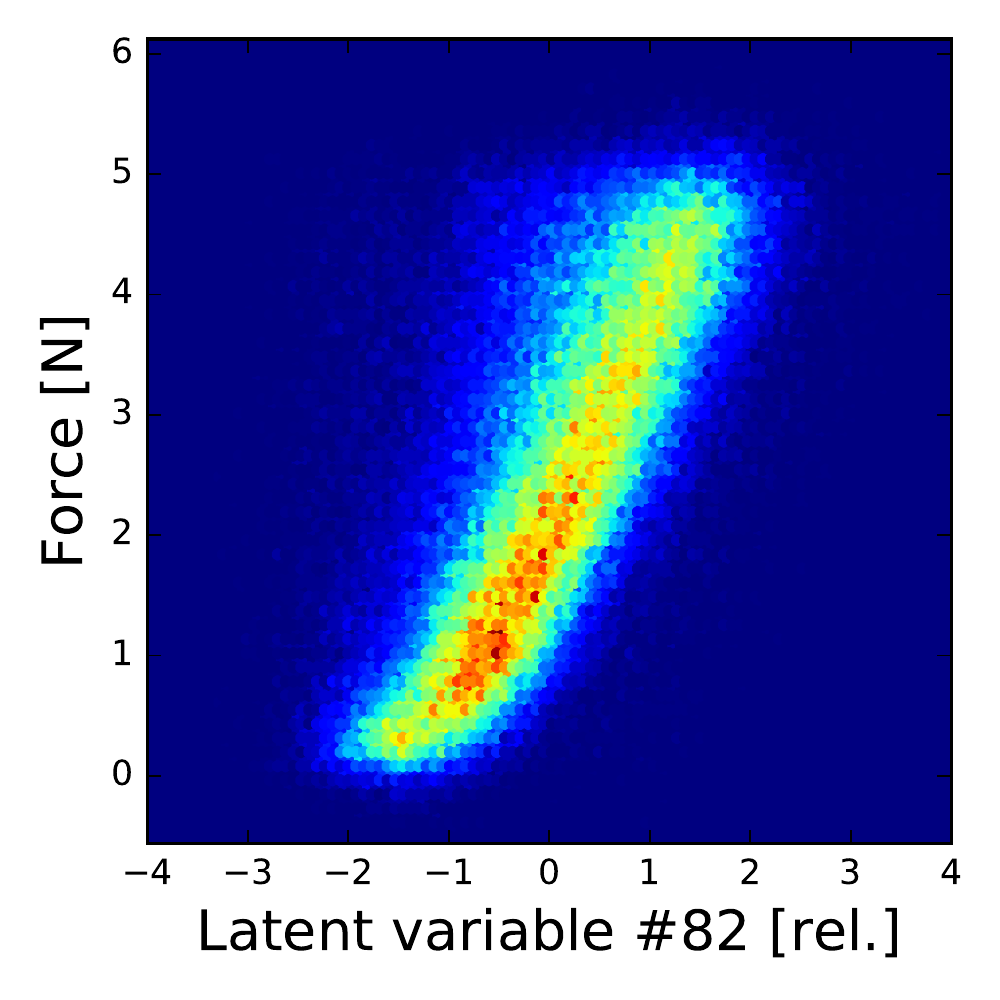} \\
      \includegraphics[width=1.5in]{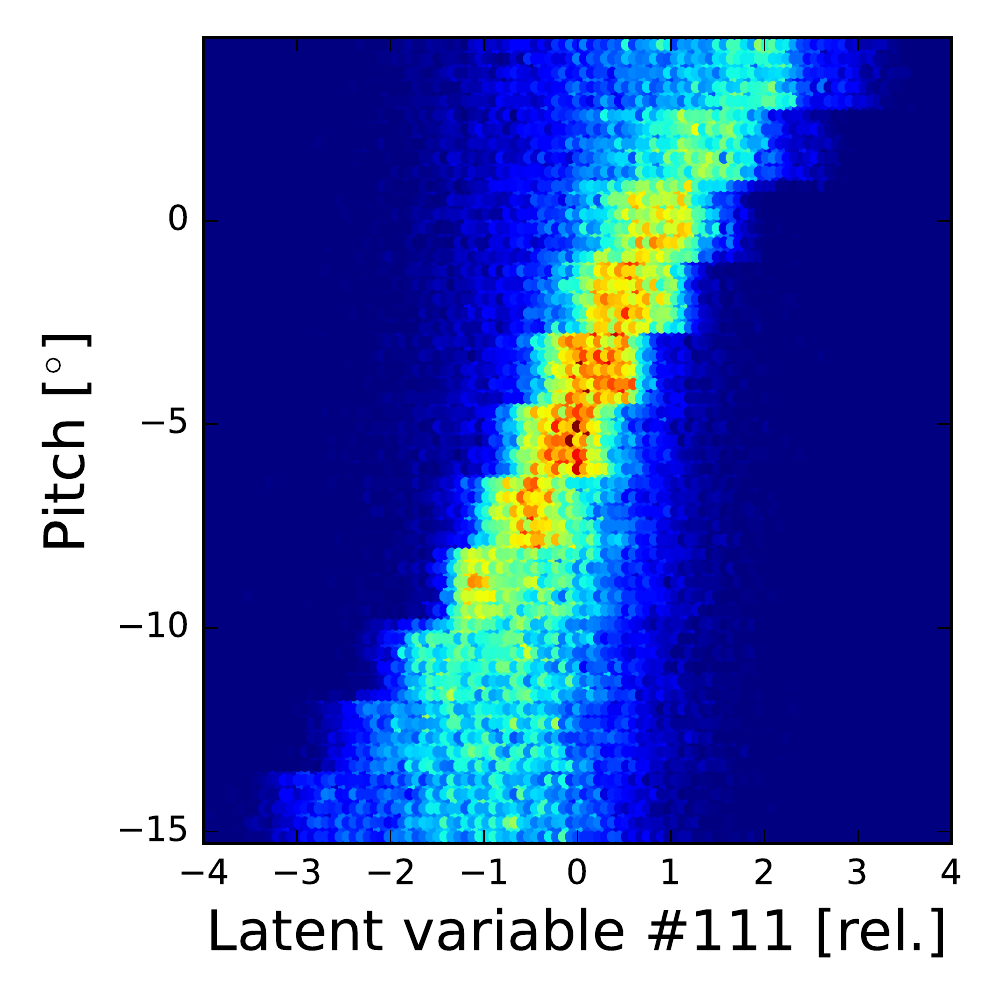} &
      \includegraphics[width=1.5in]{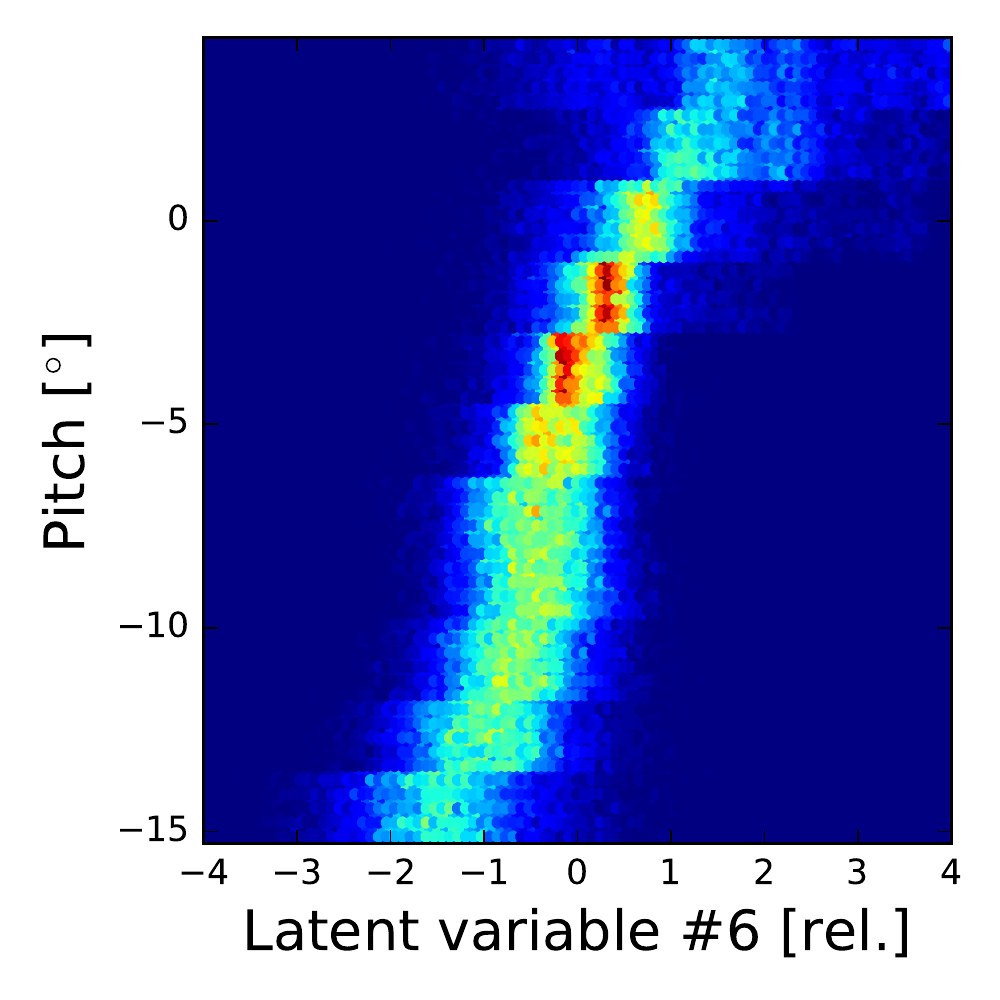} \\
      \includegraphics[width=1.5in]{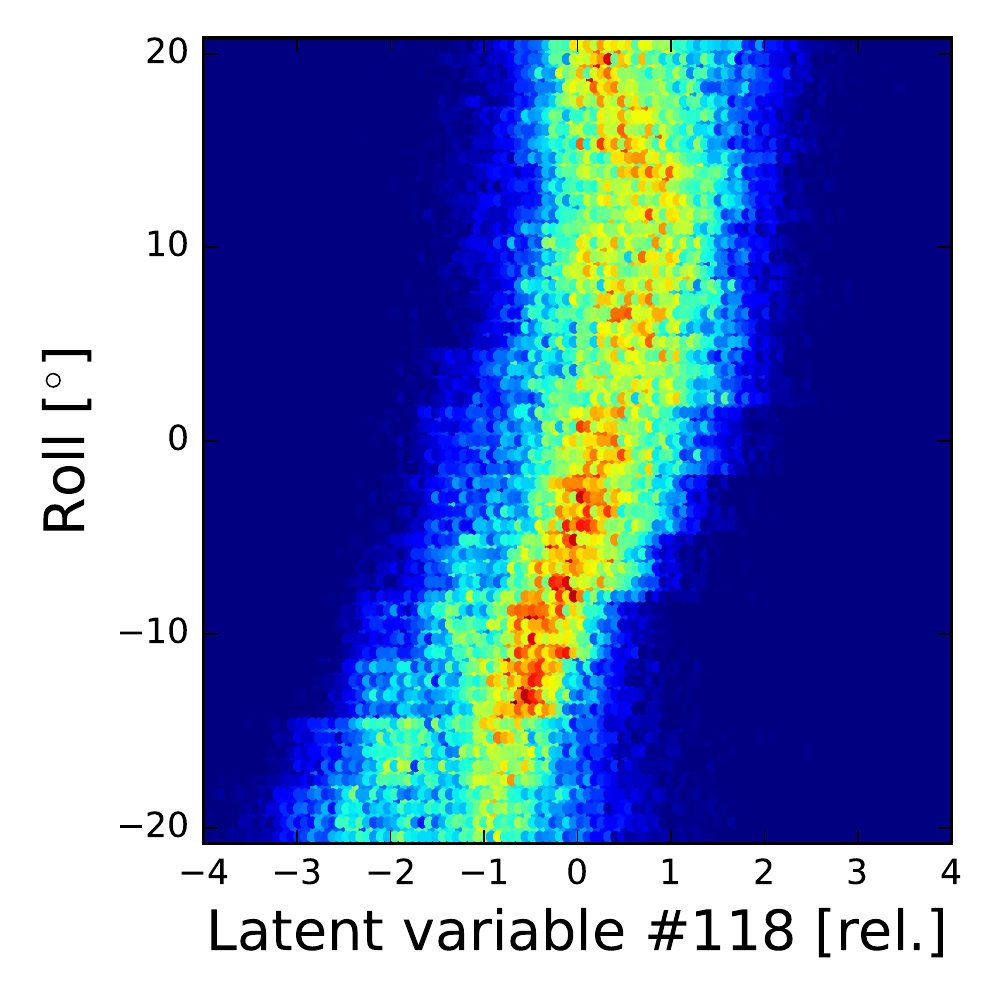} &
      \includegraphics[width=1.5in]{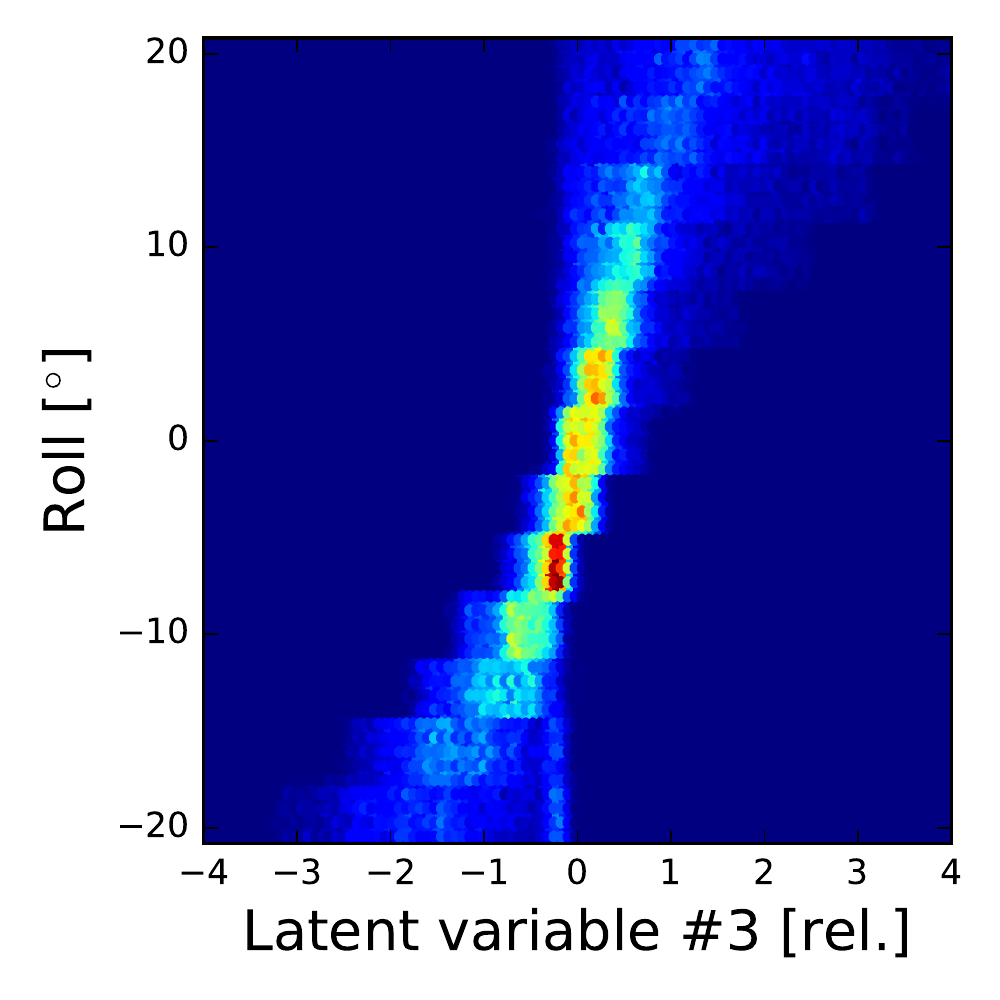}
\end{tabular}

\caption{Force and Angles from ground truth and latent variables. The Variational Auto-Encoder is able to make out underlying features of the data and stores them in individual latent nodes in a factorised format. In case of the force and surface angles dataset three difference latent nodes are each linearly related to the real physical quantities.}
\label{fig:forcelatent}
\end{figure}

We found out that the VAE algorithm we used is able to capture and separate feature in the same way as the real physical stimuli are represented: forces, angles and shore hardness are unsupervisedly learned by individual latent variables.
In Fig.~\ref{fig:forcelatent} a nearly linear relation between the latent variables with the highest correlation to the real attribute is shown.

The other elements of the latent representation correspond less to the physical value and feature a very high variance.
This makes it possible to reduce the dimension of the latent space to the minimum needed for the current dataset in an unsupervised way.
Even though the latent space is 128 elements wide, only some elements will contain information about the current tactile state. This happens due to the fact that the VAE tries to compress the data and factorises independent components of the data.
The preprocessing also manages to find the same linear relations independent of the sensor as shown in Fig.~\ref{fig:forcelatent}.
Even though both sensor are so different in their raw sensor space they now show almost the same relation between the real physical value and the corresponding latent value. 

\section{Sensor calibration}

Even though the VAE is able to unsupervisedly represent and factorise the physical quantities, it is unknown where exactly in the latent representation such quantities can be found.
This may not be hindering for control algorithms such as reinforcement learning, but can cause problems when specific features such as force are needed in a certain physical unit.
This can however be solved with a simple calibration procedure which requires only a few labelled sensor measurements, since it is only required to find the index of the element with highest correlation to the desired feature.
We recommend to rather use the full latent representation together with a suitable control algorithm since (1) the VAE also encodes tactile features which are not definable by simple physical descriptions; (2) the full resolution for specific features is only obtainable when the full latent space is used since some information is still spread among the other elements of the latent space; and (3) it completely eliminates the need to record a ground truth together with the tactile data.

\begin{algorithm}
\caption{Calibration in latent space}
\begin{algorithmic}[1]
\Require{tactile data $x$, label $z$, transformation to latent space $f(x)$}
\State{$l \leftarrow f(x)$ \Comment{transform tactile data in latent space}}
\State{$\theta \leftarrow \textrm{fit}(l, z)$ \Comment{fit linear regression from $l$ to $z$}}
\State{$i \leftarrow \textrm{argmax} \theta$ \Comment{use linear regression parameters $\theta$ to find index of latent feature}} \\
\Return{index $i$ of latent vector with highest correlation}
\end{algorithmic}
\label{alg:VAENF}
\end{algorithm}

\section{Application}

\begin{figure}
    \includegraphics[width=\linewidth,]{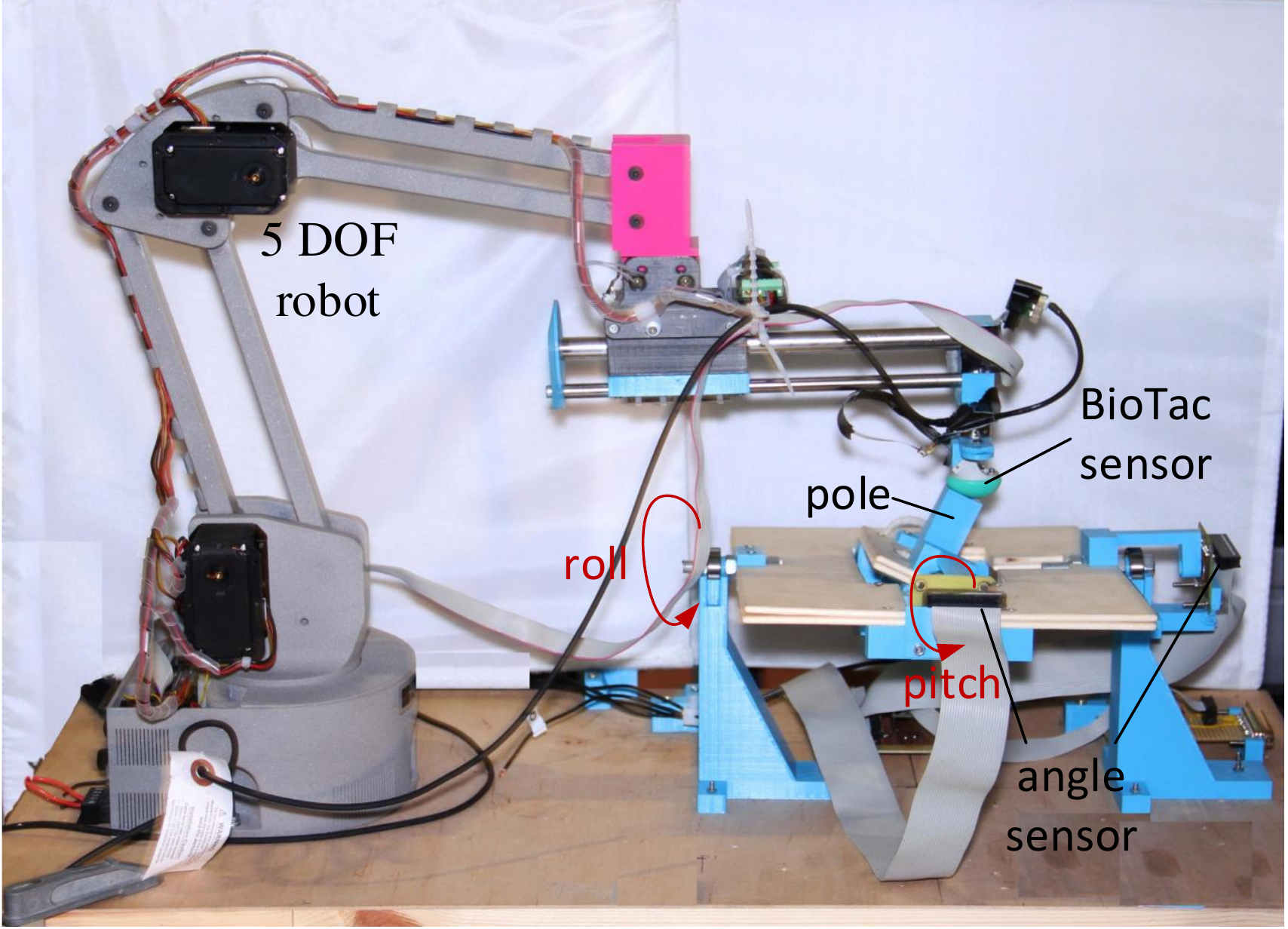}
    \caption{Experimental setup for pole stabilisation. The robot arm is touching the tip of the unstable pole mounted inside the gimbal platform}
    \label{fig:robotcontrol}
\end{figure}

We used the VAE preprocessing to stabilise a inverted pendulum using model predictive control in latent space, in order to show that the unsupervised trained features are suitable for controlling a robot.
The gimbal platform was extended with an inverted pole with a BioTac sensor touching the tip of the pole as shown in Fig.~\ref{fig:robotcontrol}.
The task was to bring the pole in an upright position.

For this control task we used a neural network to model the system dynamics as a one-step predictor.
This is done by using the state of the robot together with the current action as the neural network input and training it to predict the next state.
The network consisted of one hidden layer of 20 neurons and a rectifier activation function in each hidden unit.
We used the mean squared error as the loss for training this network.
Choosing an action is done by evaluating the neural network for all possible actions from a discrete set.
The action chosen for controlling the robot is the one where the predicted next state is the best in terms of the reward function.
We chose the reward function to be maximal at zero in the latent tactile space.
This zero position corresponded to an angle close at the centre position in angular sensor space.

\begin{figure}
\centering
\includegraphics[width=\columnwidth]{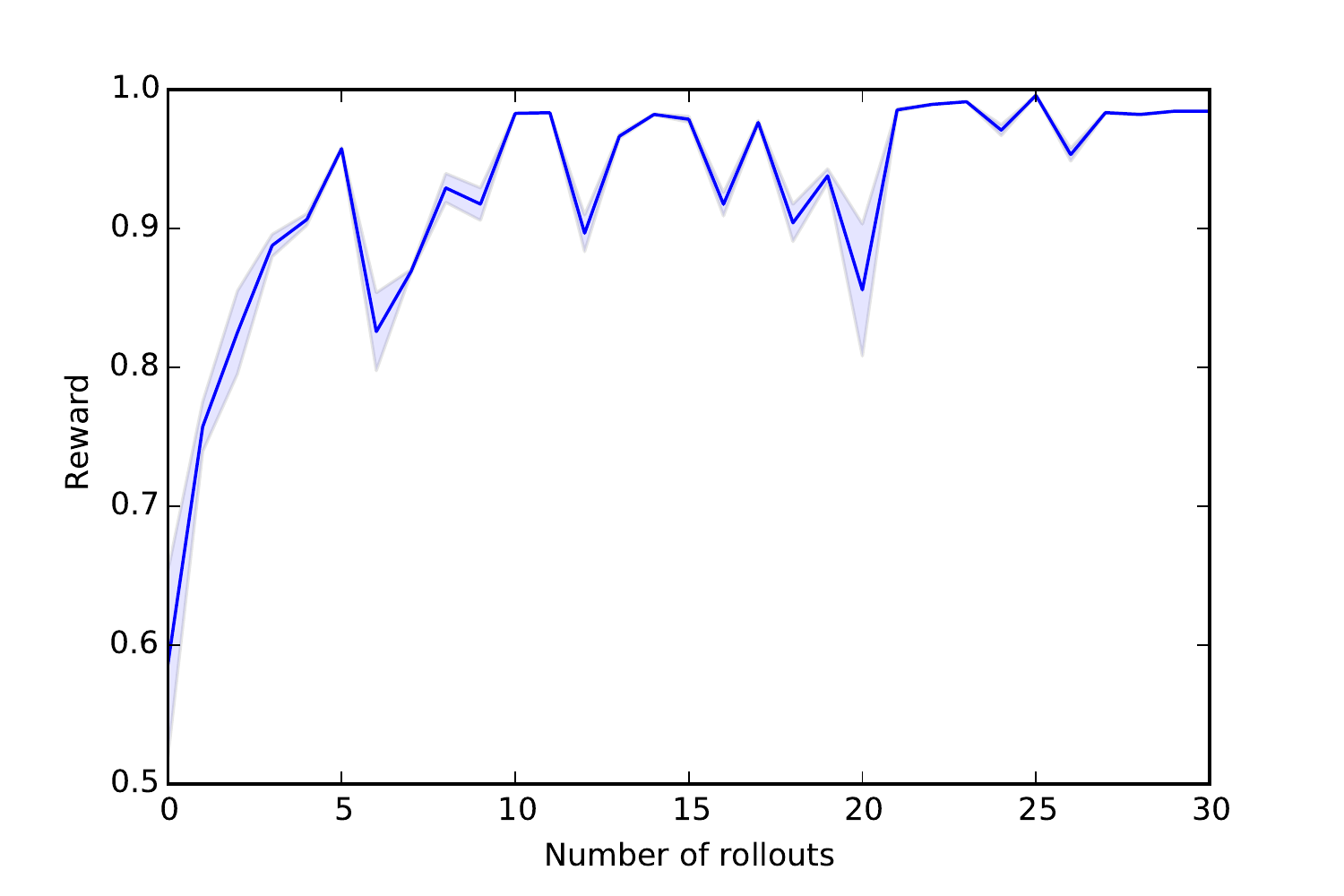}
\caption{Average reward of 10 experiments with 30 rollouts each. The controller was retrained after each rollout. The reward was calculated from last 10 time steps as each rollout starts at a random starting position.}
\label{fig:control}
\end{figure}

The results using this method can be seen in Fig.~\ref{fig:control}.
The plot shows the average reward over 10 experiments.
Training the model is performed after each of the 30 rollouts during one experiment.
As shown in Fig.~\ref{fig:control} the reward is steadily increasing until it almost reaches the maximum of 1.0 at the end of each full experiment.

\section{Conclusion}

We showed that unsupervised learning can overcome the difficult data representation that are posed by high-dimensional tactile sensors.
The preprocessing algorithm that we propose, based on the Variational Auto-Encoder, transforms the high-dimensional, sparse nonlinear tactile space into an easy-to-use compact latent space which can be directly used for control tasks.
The latent space automatically factorises the tactile features into independent components which are linearly related to real physical ground truths.
These effects can be observed in two fundamentally different tactile sensors, proving the method to be independent of the tactile sensor.
This reduces the effort to manually design or tune the preprocessing and to work completely sensor-independent.
A small control task proves the applicability of our preprocessing together with model predictive control.

\section*{Acknowledgment}

Part of this work has been supported in part by the TACMAN project, EC Grant agreement no.\ 610967, within the FP7 framework programme.

\ifCLASSOPTIONcaptionsoff
  \newpage
\fi



\bibliographystyle{IEEEtran}
\bibliography{usup}{}
%


%








\end{document}